# Systems-theoretic Safety Assessment of Robotic Telesurgical Systems

Homa Alemzadeh[1], Daniel Chen[1], Andrew Lewis[2], Zbigniew Kalbarczyk[1],
Jaishankar Raman[3], Nancy Leveson[4], and Ravishankar Iyer[1]

[1] University of Illinois at Urbana-Champaign, Urbana, IL 61801, USA
`{alemzad1,dchen8,kalbarcz,rkiyer}@illinois.edu`
[2] Applied Dexterity, Seattle, WA 98195, USA
`andrew@applieddexterity.com`
[3] Rush University Medical Center, Chicago, IL 60612, USA
`jai_raman@rush.edu`
[4] Massachusetts Institute of Technology, Cambridge, MA 02139, USA
`leveson@mit.edu`

**Abstract.** Robotic surgical systems are among the most complex medical cyber-physical systems on the market. Despite significant improvements in design of those systems through the years, there have been ongoing occurrences of safety incidents that negatively impact patients during procedures. This paper presents an approach for systems-theoretic safety assessment of robotic telesurgical systems using software-implemented fault-injection. We used a systems-theoretic hazard analysis technique (STPA) to identify the potential safety hazard scenarios and their contributing causes in RAVEN II, an open-source telerobotic surgical platform. We integrated the robot control software with a software-implemented fault-injection engine that measures the resilience of system to the identified hazard scenarios by automatically inserting faults into different parts of the software. Representative hazard scenarios from real robotic surgery incidents reported to the U.S. Food and Drug Administration (FDA) MAUDE database were used to demonstrate the feasibility of the proposed approach for safety-based design of robotic telesurgical systems.

## 1 Introduction

In an analysis of over 10,000 adverse events related to robotic surgical systems, reported between 2000–2013 to the U.S. FDA MAUDE database [1], we showed that 9,382 (88.3%) of the reported events involved device and instrument malfunctions. Those events had significant negative patient impacts, occasionally leading to deaths and injuries or causing procedure interruptions to troubleshoot system problems. In particular, out of 536 system errors detected during procedures, 488 (91%) led the surgical teams to manually restart the system (43% of 488), convert the procedure (61.5%), or reschedule it to a later date (24.8%). (Note that these categories are not mutually exclusive. In some cases after several system resets, the procedure was converted or rescheduled.) [2]. This data shows the importance of designing robust safety features in robotic surgical systems and verifying the effectiveness of detection and recovery mechanisms in order to prevent similar safety hazards in the future.

The international safety standards (e.g. ISO 14971 for medical devices and ISO 26262 for automobiles) recommend identifying potential safety hazards and defining safety requirements to implement mechanisms that can detect and mitigate hazards. The standards also emphasize the importance of fault-injection testing as a means to validate the robustness of safety mechanisms in presence of faults and abnormal conditions [3].

However, traditional hazard analysis techniques primarily focus on the failures of individual components or human errors in the system. Other potential causal factors, such as complex software errors and unsafe component interactions, are often not thoroughly considered during the analysis. Systems-theoretic hazard analysis techniques such as STPA (Systems-Theoretic Process Analysis) [4] overcome this limitation by modeling accidents as complex dynamic processes resulting from inadequate control mechanisms that violate safety constraints. It is shown that STPA can identify additional causes for accidents that are not detected by FTA and FMEA techniques [4, 5].

Software implemented fault injection (SWIFI) [6, 7] is a common technique for validating the effectiveness of fault-tolerance mechanisms by studying the behavior of the system in presence of faults. The effects of software or hardware faults are emulated by randomly changing code or data at different software locations. However, with the increasing size of software in today's complex systems, it is a challenging task to define specific fault types and locations that can effectively emulate realistic fault scenarios.

In this work, we took a systems-theoretic approach to empirically validate the robustness of safety mechanisms in robotic telesurgical systems by identifying the critical locations within the system to target software fault-injection. More specifically, we used the potential causes of safety violations identified by STPA to define types and locations of faults to be injected in robot control software in order to evaluate the system under realistic hazard scenarios. As a case study, we used RAVEN II robot, an open-source platform for research in telerobotic surgery [8]. We developed a software fault-injection framework that mimics the control flaws identified during hazard analysis by automatically injecting faults into robot control software modules. We evaluated the feasibility of the proposed approach using examples of real adverse events from the FDA MAUDE database, which resemble the hazard scenarios identified in our analysis.

## 2    Background

### 2.1    RAVEN II Robotic Surgical Platform

The RAVEN II robot is an open-source platform for research in tele-operative robotic surgery [8]. Fig. 1 depicts a typical configuration of a robotic telesurgery system, composed of a master console, communication channel, and a RAVEN II surgical robot, including software and hardware components. The **master console** provides the means for the surgeon to issue commands to the robot using foot pedals and master tool manipulators. The desired position and orientation of robotic arms, foot pedal status, and robot control mode are transferred between the master and slave robot over the network using the Interoperable Teleoperation Protocol (ITP), a protocol based on the UDP packet structure [9]. RAVEN II **control software** receives the user command packets, translates them into motor commands, and sends them to the **control hardware**, which enables the movement of robotic arms and instruments.

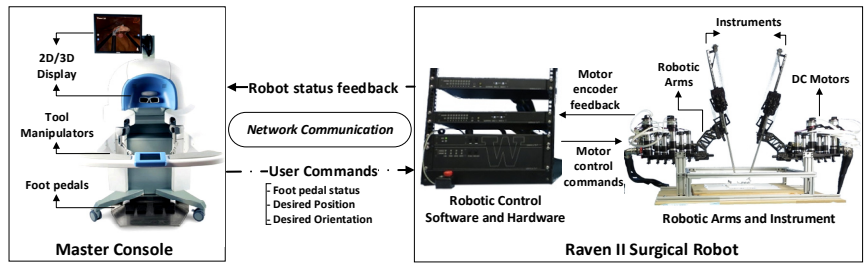

**Fig. 1.** Robotic Telesurgery using RAVEN II Surgical Platform (modified from [10, 11])

Fig. 2 shows the main hardware and software modules in the RAVEN II control system. The control software runs on top of the Robotic Operating System (ROS) middleware and real-time (RT-Preempt) Linux kernel and communicates with the motor controllers and a Programmable Logic Controller (PLC) through custom USB interface boards. The PLC controls the brakes. The motor controllers send movement commands to the motors. There are three main threads running in parallel in the RAVEN control software: 1) the **network layer** thread which receives the command packets from the master controller over network; 2) the **control thread** where the robot's kinematics and control computations are performed; and 3) the **console thread** which provides an interface for setting the control modes and displaying robot's status to user.

Both the control software and the PLC operate in a state machine that consists of four states: a) emergency stop ("E-STOP"), b) initialization ("Init"), c) foot pedal released ("Pedal Up"), and d) foot pedal pressed ("Pedal Down"), as shown in the Fig. 2.b. The control software's state is synced with the PLC state every 1 milliseconds through the USB interface boards. At power-up, both software and PLC are at "E-STOP" state, the motor brakes are engaged, and motor-controllers are stopped. As a safety mechanism, the robot has a physical start button which should be pressed in order to start the robot initialization (homing) process and make it ready for manual teleoperation. The initialization state takes each robotic arm from its resting position and moves it into the surgical field. Once the homing process is done the system automatically transitions into the "Pedal Up" state where the brakes are engaged and robot does not move. The "Pedal Down" state is initiated when the human operator pushes the foot pedal down. In this state the brakes are released, allowing the master console to directly control the robot. When the human operator lifts their foot from the pedal, the system re-enters "Pedal Up" state, disengaging the master console from the robot. There is an emergency stop button that immediately stops the robot by putting the PLC safety processor and the control software into "E-STOP" state [12].

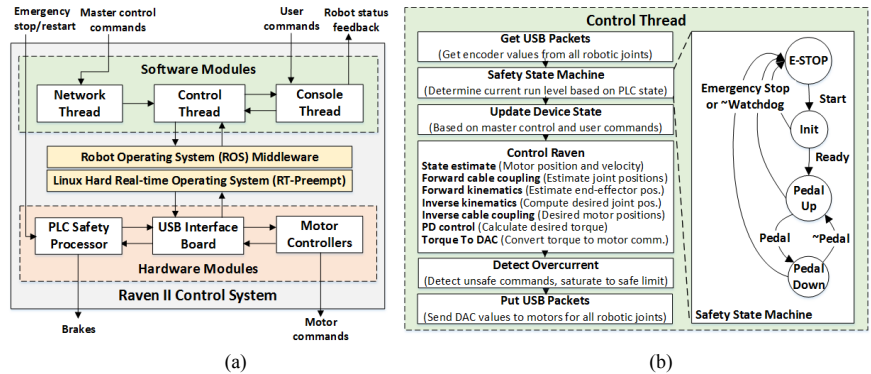

**Fig. 2.** RAVEN II control system: (a) Software and hardware modules [8, 11], (b) Computation steps in the control thread when in the Cartesian space mode and the safety state machine [11]

The control software detects and corrects any unsafe motor commands (e.g., electrical currents directed to the motor controllers exceeding a safe limit) using *overdriveDetect* function. During normal operation, the software continuously sends a square-wave watchdog signal to the PLC through the USB boards. Upon detecting an instant over-current command by *overdriveDetect* function, the control software stops sending the watchdog signal. The watchdog timer implemented in the PLC safety processor monitors the periodic watchdog signal from the software and upon loss of the signal immediately puts the system into the emergency stop ("E-STOP") state.

### 2.2 Systems-theoretic Hazard Analysis Using STPA

STPA is a hazard analysis technique based on STAMP (Systems-Theoretic Accident Model and Processes) accident causality model which is driven by concepts in systems and control theories [4]. STAMP models the systems as hierarchical control structures, where the components at each level of the hierarchy impose safety constraints on the activity of the levels below, and communicate their conditions and behavior to the levels above. The interactions among system components and operators are modeled as control loops composed of the actions or commands (e.g., motor commands) that a controller (e.g., software controller) takes/sends to a controlled process (e.g. the robotic arms/instruments) and the response or feedback (e.g., joint positions) that the controller receives from the controlled process (see Figure 3.a). The layers of the control structure could span from the physical components to human operators, up to even higher levels in manufacturing, management, and regulation. In every control loop, the controller uses an algorithm to generate the control actions based on a model of the current state of the process it is controlling (see Fig. 3.b). The control actions taken by the controller change the state of the controlled process (e.g., the instrument will be engaged). The feedback (e.g., motor encoder values) sent back from the controlled process (e.g., motor controllers) update the process model used (e.g. current joint status) by the controller.

STPA starts by defining the accidents to be considered, the hazards associated with those accidents, and the safety requirements (constraints) for the system. Then unsafe control actions in each loop of control structure are identified and the potential causes for unsafe controls are determined by considering any potential flaws in the inputs, control algorithm, process model, outputs, or feedback received by the controller [4].

## 3 Systems-theoretic Safety Validation Using Fault Injection

We used systems-theoretic hazard analysis using STPA to identify the safety hazards of a typical robotic telesurgical system and the potential causal factors that might lead to safety violation in RAVEN II system. Then we validated the robustness of the safety mechanisms of RAVEN II to the safety hazard scenarios identified using STPA by simulating their causal factors using software-implemented fault-injection.

### 3.1 Safety Hazards and Unsafe Control Actions

First, based on the review of almost 1,500 accident reports from the FDA MAUDE database and specification of RAVEN II system functionality, we classified the accidents in robotic surgical systems into three types: patient deaths (A-1); patient injuries during procedure or serious complications experienced after procedure (A-2); and costly damage to surgical system or instruments (A-3). We also identified three main system hazards or set of system conditions that could lead to these accidents (Table 1).

Table 1. Accidents and safety hazards in robotic surgical systems

| Accidents |
|---|
| **A-1.** Patient expires during or after the procedure. |
| **A-2.** Patient is injured during procedure or experiences complications after the procedure. |
| **A-3.** Surgical system or instruments are damaged or lost. |
| **System Hazards** |
| **H-1.** Robot arms/instruments move to an unintended location (H1-1), or with an unintended velocity (H1-2), or at unintended time (H1-3). |
| **H-2.** Robotic arms or instruments are subjected to collision or unintended stress. |
| **H-3.** Robotic system becomes unavailable or unresponsive during procedure. |

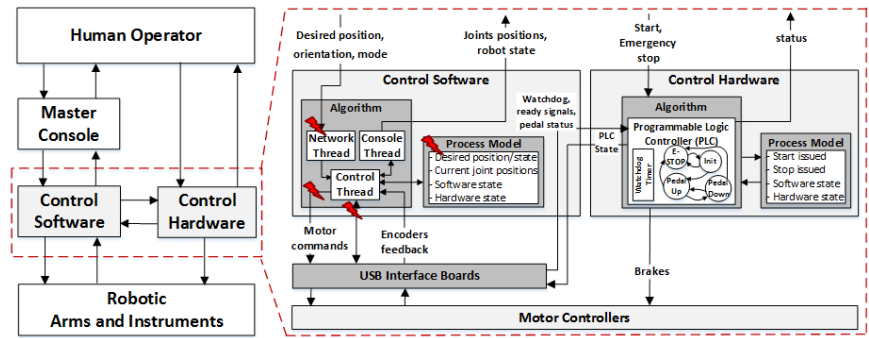

**Fig. 3.** a) Hierarchical control structure of RAVEN II system, b) Software and hardware control loops, including control algorithms, process models, control actions, and feedback.

We then modeled the hierarchical safety control structure of the system, as shown in Figure 3. Software and hardware control loops (outlined by dashed lines in Fig. 3.a) are further refined in Fig. 3.b to illustrate details of the interactions among the software and hardware controllers. Next, we identified the set of system conditions under which the control actions could possibly be unsafe and lead to hazardous system states. The following unsafe scenarios were specifically considered: i) a required control action was not performed, ii) a control action was performed in a wrong state, leading to a hazard, iii) a control action was performed at an incorrect time, iv) a control action was performed for an incorrect duration, v) a control action was provided, but not followed by the controlled process [4]. For example, in software control loop shown in Fig. 3.b any flaws (marked with ⚡) in the master console inputs, the incorrect feedback from the motor controllers or hardware control, and the flaws in the process model of software or output generated by the control algorithm can be considered as a potential causal factor. Table 2 shows the potentially unsafe control actions and their corresponding possible causal factors for the software and hardware control loops.

As shown in the next section, the identified causal factors in combination with the knowledge of software structure provide the scope for performing directed fault-injection experiments. They can define the location within each software module to inject, the variables within each function to target, and the conditions to trigger the injections.

**Table 2.** Potential unsafe control actions and causal factors for safety hazards in RAVEN II

| Control | Control Action (Type) | Context (System Condition) | Safety Hazards | Possible Causal Factors |
|---|---|---|---|---|
| Software Control | Motor command (provided) | User desired joint position does not match user desired position | H1-1 | - Incorrect console inputs<br>- Faulty control algorithm<br>- Incorrect process model (desired positions, joint positions, runlevel)<br>- Faulty USB communication<br>- Arms/Instruments malfunctions |
| | | User desired joint position is at a large distance from the current joint position (unintended jump) | H1-2 | |
| | | Left and right arm end-effector positions are very close (unintended collision) | H2 | |
| | | Software State = E-STOP or Software State = Pedal Up, PLC State = Pedal Down | H1 H2 | - Missing/incorrect input from PLC<br>- Faulty control algorithm<br>- Incorrect process model (desired positions, joint positions, runlevel)<br>- Missing/incorrect watchdog signal or output to PLC<br>- Faulty USB communication |
| | | Software State = Pedal Down, PLC State = Pedal Up or PLC State = Init | H3 | |
| | | Software State = Not E-STOP, PLC State = E-STOP | H3 | |
| | Motor command (not followed) | Software State = Pedal Down or Software State = Init | H3 | - Faulty USB communication<br>- Mechanical malfunctions (e.g. broken instruments or cables) |
| Hardware Control | Brake (provided) | Stop not pressed and Software not stopped/pedal up | H3 | - Missing/incorrect watchdog signal or output from software<br>- Faulty USB communication |
| | Brake (not provided) | Stop pressed or Software is stopped | H1 H2 | |
| | Brake (not followed) | | H1 H2 | - Mechanical malfunctions (e.g. broken instruments or cables) |

### 3.2 Safety Hazard Injection Framework

To evaluate the safety mechanisms of robotic surgical system, we developed a Safety Hazard Injection Framework, which consists of seven modules for retrieving hazard scenarios, generating fault injection campaign, selecting fault injection strategy, conducting fault injection experiment, and collecting data, all in an automated fashion. Fig. 4 shows the overall architecture of these modules and how they interface with each other and with the RAVEN II control software and hardware. A detailed description of each module is provided below.

**Injection Controller.** The Injection Controller is responsible for starting, stopping and automating the fault injection campaign. It communicates with other modules in the Safety Hazard Injection Framework through sockets or by direct invocation. In a normal campaign execution, it first accesses the Safety Hazard Scenario Library to retrieve the list of hazard scenarios. Second, the controller calls the Fault-Injection Strategies to generate the fault injection parameters that could cause each hazard scenario. Next, it runs the user input generator module and calls the appropriate Fault-Injector and robotic software to conduct a fault injection experiment. At the end of each injection run, the injection parameters and data are collected and written to the Data Collector.

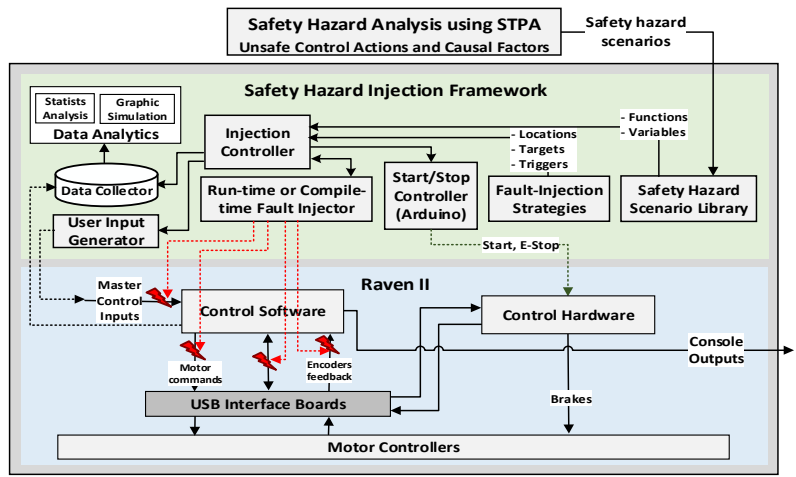

**Fig. 4.** Safety Hazard Injection Framework integrated with the RAVEN II Surgical Platform

**Safety Hazard Scenario Library.** The safety Hazard Scenario Library contains the safety hazard scenarios identified during the hazard analysis using STPA. Each hazard scenario includes a possible unsafe control action that might happen in the system and a list of potential causal factors. An example unsafe control action would be a motor command is provided by the control software when *there is a mismatch between the software state and hardware state of the robot* (rows 4-6 in Table 2). *Faulty USB communication* is an example causal factor that might lead to such unsafe control action.

**Fault-Injection Strategies.** Based on the causal factors involved in each safety hazard scenario, the analysis of RAVEN source code, and software/hardware architecture, the Fault-Injection Strategies module retrieves information on software functions which can most likely result in the hazard scenario, as well as the key variables in those functions and their normal operating ranges. This information is translated to the parameters to be used by the fault-injection engine for simulating potential causal factors and validating whether they lead to the unsafe control or the safety hazards in the system. The fault injection parameters include the *location* in the software function, the *trigger* or condition under which the fault should be injected and the *target* variables to be modified by the injection.

**User Input Generator.** User input generator emulates master console functionality by generating input packets based on a previously collected trajectory of robotic movements made by a human operator and sends them to the RAVEN II control software.

**Fault Injectors.** The Fault Injectors perform the fault injection during robot operation. We developed both compile-time and run-time fault injectors with minimum changes to the RAVEN software and hardware. Run-time fault injector is implemented by extending the functionality of GDB (GNU Project Debugger for Linux). More specifically, we extend the *breakpoint* feature in GDB to perform fault injection when the desired *trigger* condition is met and then resume the execution of the target program. Run-time fault-injector launches the RAVEN ROS node with GDB Server attached to it, then the extended GDB is run from a remote process and after connecting to the RAVEN node, performs the fault injections. Run-time fault injector has the advantage of performing injections on run-time generated data; however the delay introduced by the run-time breakpoints is not acceptable for modules that have hard real-time requirements. For example, the RAVEN control thread has the hard real-time requirement of one millisecond to perform kinematics calculations and communication with the USB boards [11]. Run-time fault-injection to the control thread introduces small delays, leading to violation of the real-time constraint and failure of kinematics calculations, resulting in unintended robotic instrument vibrations and movements. Compile-time fault injector is implemented as a module that modifies and recompiles the fault injection conditions into the source code. The main advantage is negligible timing overhead (small compile and build times), which is acceptable for modules with hard real-time requirements. We use compile-time injector to inject faults into the control thread.

**Start/Stop Controller.** To perform automated fault-injection experiments without manual user intervention, we added a hardware mechanism to automatically start and stop the RAVEN system. We connected the start input of the PLC to the output of a relay switch controlled by an Arduino microcontroller (http://www.arduino.cc/) which receives start signals from the Injection Controller. After each injection, the controller stops the system by shutting down the RAVEN ROS node. The next injection gets started by automatically launching the software and sending the start signal to the Arduino relay controller to start the PLC and homing process.

**Data Collection and Analysis.** For each fault injection run, the fault injection parameters, surgical robot's trajectory, and detected errors are collected and sent to a MySQL server on a remote machine (Data Collector). These data are later queried for statistical analysis or graphics simulation.

9## 4 Experimental Results

In our experiments, we simulated 45 scenarios (corresponding to the causal factors shown in Table 2) by injecting faults into 25 locations within 13 software functions of the network and control threads of RAVEN II robot, while running a pre-collected trajectory of a surgical movement. We ran a total of 2,146 fault-injection experiments on the RAVEN robot. However, the majority of the faults (e.g., injected values within the range of variables) were not manifested in the system, or their effect was not logged completely by the data collection process due to system hangs/crashes (e.g., hardware "E-STOP") caused by the faults. Table 3 shows examples of scenarios where the faults were manifested in the system. For each scenario, we conducted multiple runs (in total 368 fault injections) to get confidence in reproducibility of the manifested/observed system behavior and manually collected the results. In each case we analyzed the system behavior both during the homing process (which system is being initialized and user manipulation has not started yet) and after the homing. The third column in the table shows the number of experiments done for each scenario and the last column corresponds to the scenario ID. A complete list of causal scenarios is available at [13].

In this section, we discuss our findings from the conducted fault injection experiments, including the causes for undetected hazards and the hazard scenarios that were mitigated by the safety mechanisms. Next section shows representative incident reports from the MAUDE database, which resemble the safety hazard scenarios identified here.

### 4.1 Undetected Safety Hazards

In what follows we describe the scenarios in which the injected faults led to hazards that were not detected or mitigated by the safety mechanisms in the system.

**Unintended Robotic Movement (H1).** We found a total of six scenarios where the faults in the console inputs, control algorithm, or the communication between the control software and hardware led to robotic arms/instruments making movements to an unintended position (H1-1) or with an unintended velocity (H1-2).

i. Out of range values injected permanently into the position, orientation, and foot pedal status inputs received from the master console (in *network_process* function) did not have any impact on the system during the homing process. However, after homing and in "Pedal Down" state, these injections led to kinematics calculations failures, small jumps, or stopping the robot. If the injected values passed the safe limits, movement was stopped by the overdrive detector and E-STOP was raised.

ii. Intermittent injection of out-of-range values into the master console inputs occasionally caused small instrument jumps or stopping the PLC when the faults were injected at very high frequency (e.g., at every other cycle).

iii. Injecting a random constant torque value to the joints current commands sent from the control software to the motor controllers (in *TorqueToDAC* function) caused very abrupt jumps of robotic arms, which resulted in the breakage of cables on the arms.

iv. Faulty estimation of motor velocities by the control algorithm (in *stateEstimate* function) caused unintended rotation and movement of instruments. In one case, upon intermittent injection of zero velocity, the instruments unexpectedly overshot the home position and collided with the surgical field floor during homing process.

v. Intermittent faulty packets received by the USB interface function (*getUSBPacket*) from the PLC caused the software control to assume that PLC is in "E-STOP" state, while PLC was in "Init" state. During homing process, this fault led the software and PLC to switch back and forth from "Init" to "E-STOP" state, causing failure of synchronization between left and right arms. Therefore, the robotic system got stuck in the initialization process and never moved to "Pedal Up". After homing, depending on the frequency of the intermittent faults, either the robot completely stopped or PLC applied brakes repeatedly to the motors.

vi. Injecting faults into the packets sent to the motor controllers through the USB interface function (*putUSBPacket*) did not impact the behavior of the system during the homing process, but led to abrupt jumps of robotic arms, resulting in cable breaks. A video recording of this scenario is available at [13].

**Unintended Collision or Mechanical Stress (H2).** The last four scenarios discussed above (iii - vi) also involved mechanical stress on the robot due to hanging in the homing process, repeating initialization steps, applying brakes over and over again, abrupt jumps of robotic arms, colliding with the surgical field floor, or breaking cables. The robotic system also became unresponsive or unavailable (H3) for almost an hour while repairing each broken cable. Due to the risk of damage to the robot, we repeated these specific injections only a few times.

**Unresponsive Robotic System (H3).** The majority of undetected safety hazards were due to faults injected in the USB communication or communication between software and PLC (17 scenarios [13]), leading the robotic system to not start the homing process, stop movement, become unresponsive to the received console packets, or become unavailable due to mechanical issues. Table 3 shows examples of these scenarios (vii, viii).

In case of transient or intermittent faults (e.g., in input console packets or USB packets), restarting the system can resolve the E-STOP conditions. However, permanent faults (e.g., a loose or disconnected USB cable causing incorrect information sent from PLC to software, or a DAC malfunction causing incorrect values sent to the motors, simulated as stuck at software faults here) cannot be recovered from even after multiple restarts and by hanging in E-STOP state the robotic system becomes unavailable (H3).

### 4.2 Mitigated Safety Hazards

Out of 23 scenarios related to corruption of the console inputs and the control algorithm, only 6 caused the unintended movements (depending on the robot configuration), collision, or cable damage. All these cases where related to intermittent faults (out of range absolute values) injected into the console inputs (tool positions and orientation or foot pedal) in a periodic manner or to applying constant velocity/torque values to the motors. All other scenarios either did not have any impact (3 cases), were detected by the *overdriveDetect* function and mitigated by forcing a hardware "E-STOP" (9 cases) (see scenario ix in Table 3), or only caused the system to hang in "Pedal Up" or "E-STOP" with no potential harm (4 cases).

11**Table 3.** Example scenarios simulated by fault injection and the observed system behavior

| Potential Causal Factor | Injected Software Fault *Target Function*: Variables [Fault Type, Values] | No. | Observed System Behavior | Hazard | ID |
|---|---|---|---|---|---|
| Incorrect console inputs | *network_process*: Position and Orientations [Stuck At Out of Range] | 20 | During Homing: No impact<br>After Homing in Pedal Down: IK-failure, small jumps, no movements with no E-STOP, E-STOP | H1 H3 | i |
| | *network_process*: Foot Pedal Status [Stuck At 0, StuckAt 1] | 20 | During Homing: No impact<br>After Homing: Does not start movement if Stuck At 0, No impact if Stuck at 1. | | |
| | *network_process*: Position and Orientations [Intermittent Out of Range every10, 100, 500 packets] | 40 | Homing: No impact<br>After Homing in Pedal Down: IK-failure, No movement, small jumps with no E-STOP, or E-STOP depending on robot configuration | H1 H3 | ii |
| | *network_process*: Foot Pedal Status [Intermittent 0/1 Flip every 30,100,3000 cycles] | 20 | Pedal Down: Movement stops or small jumps<br>PLC stops at very high flipping rate (e.g. every other cycle) | | |
| Faulty control algorithm | *TorqueToDAC*: Joints Current Commands [Stuck At -1000] | 1 | Abrupt jump of both robotic arms,<br>Cables on both left and right arms broke | H1 H2 H3 | iii |
| | *stateEstimate:* Motor Velocity [Stuck At 0, -1, 1000] | 5 | During Homing: Unintended rotation, E-STOP<br>After Homing: No Impact | H2 | iv |
| | *stateEstimate:* Motor Velocity [Intermittent 0 injection every 100, 3000 cycles] | 5 | During Homing: Unintended tool movement, hard collision of instrument to the floor<br>After Homing in Pedal Down: No impact | H1 H2 H3 | |
| | *stateEstimate:* Motor Position [Stuck At or Intermittent] | 10 | Detected and mitigated by (*overdriveDetect*) Raised E-STOP Error and Stopped | NA | ix |
| Faulty USB communication | *getUSBPacket:* PLC State [Stuck At 0] | 12 | Homing: Does not start initialization, software assumes hardware is in E-STOP<br>After Homing: E-STOP, software assumes hardware is in E-STOP, goes to E-STOP, stops sending watchdog, causing hardware to really stop. | H3 | vii |
| | *getUSBPacket:* PLC State [Intermittent 0 injection] | 10 | Homing: Repeats the homing process over and over again due to synchronization failure between two arms.<br>After Homing: Hardware completely stops or brakes are engaged/disengaged repeatedly | H2 H3 | v |
| | *putUSBPacket:* Joints Current Commands [Stuck At Random Value] | 5 | During Homing: No Impact.<br>After Homing: Abrupt jump of robotic arms and cable breaks, Software E-STOP | H1 H2 H3 | vi |
| Incorrect output to PLC | *updateAtmelOutputs:* Output to PLC [Stuck At 0, 1, 3] | 16 | Does not start the initialization process or stops after homing because hardware goes to E-STOP and gets stuck there | H3 | viii |

To appear in the International Conference on Computer Safety, Reliability, and Security (SAFECOMP)
Copyright © 2015: Authors.

## 5 Discussion

### 5.1 Related Safety Incidents from FDA MAUDE Database

Table 4 shows representative incident reports from the FDA MAUDE database, related to the da Vinci surgical system (the only surgical robot for minimally invasive surgery available on the market) [14]. In these examples, similar hazard scenarios studied in this paper (including master console malfunctions and communication failures between the controller and robotic parts) occurred during real robotic procedures. These failures led either to non-intuitive movement of instruments or system errors that could not be cleared even by multiple system restarts.

**Table 4.** Relevant incident reports on da Vinci surgical system from FDA MAUDE database

| Report # (Year) | Summary Event Description from the Report | Potential Causal Factors (ID in Table 3) | Observed Behavior (Hazard) | Patient Impact |
|---|---|---|---|---|
| 2120175 (2011) | During a hysterectomy procedure, the left master controller did not have full control of the maryland bipolar forceps instrument, resulting in non-intuitive motion and causing a small bleed on the patient's uterine tube. | Master console calibration issue (i) | Non-intuitive movement (H2) | Small bleed on patient's uterine tube |
| 2663924 (2012) 2589307 (2012) | Approximately 3.5 hours into a pancreatectomy procedure, multiple instances of non-recoverable system error code #23 was experienced and the surgeon was unable to control the patient side manipulator (psm) arms. | Communication failure between master console and robot (i) | Non-recoverable system error (H3) | Converted to open surgery after 3.5 hours |

In cases of instruments moving of their own accord or getting stuck due to malfunctions, the consequences may range from minor, where there are just short delays or system resets for troubleshooting the problem, to major, where the instruments may impale or impinge on a bodily structure, causing perforation or bleeding. Tearing or perforation of tissues can cause long term complications and even death. Conversion of procedure to non-robotic approaches is a recovery mechanism to ensure survival of the patients. However, lack of tactile feedback can be a major issue in extracting malfunctioning instruments safely from patient's body.

This study demonstrated the value of software-implemented fault injection for simulation of safety hazard scenarios, which might help surgeons recognize complications and act promptly to prevent similar incidents in the future.

### 5.2 Vulnerabilities in Safety Mechanisms and Mitigation of Safety Hazards.

We discovered the following vulnerabilities in the safety mechanisms of RAVEN II robot which contributed to the simulated safety hazards:
  a) Lack of monitoring mechanisms for the initialization (homing) process.
  b) No safety mechanisms for monitoring the USB board communications.
  c) No hardware detection mechanisms for monitoring unsafe motor commands.
  d) No feedback from the motor controllers and brakes to the PLC

The initial specifications of the RAVEN robot [12] included the requirements for the PLC to monitor the robotic hardware through feedback received from the motors and brakes. However, we found that those monitoring mechanisms are not included in the current implementation of the robot. Also, separate software and hardware mechanisms for monitoring the activities of USB interface boards are needed in the future.

The following robust safety mechanisms had a major role in mitigating safety hazards in RAVEN II, by preventing unintended movements and possible system damage:
  a) Robot movements cannot start without a start signal provided by the user.
  b) PLC engages the brakes upon loss of watchdog ("E-STOP") or foot pedal signals from software ("Pedal Up"); and software only sends the pedal signal to the PLC when the foot pedal is pressed and it is not in "E-STOP" or "Init" state.
  c) Software checks the status of PLC on every cycle (1 millisecond interval) to immediately follow the state transitions of the robotic hardware.

## 6   Related Work

Software implemented fault injection (SWIFI) [6, 7] has been used for evaluating the dependability of different computing systems, including operating systems [15], smart power grids [16], and SaS cloud platforms [17]. International safety standards, such as NASA Software Safety Guidebook and functional safety standard for automobiles (ISO 26262), recommend using fault-injection for validation of safety-critical software [3]. However, medical devices safety standard (ISO 14971) do not consider fault-injection testing for validation of medical software [18]. Only one study showed the use of software simulation fault injection for testing the UML model of software for a pacemaker [19]. In this work, we developed a software fault injection framework that targets the critical locations in a real medical cyber-physical system to validate the robustness of the system safety mechanisms during design and implementation phases.

STPA was previously used for hazard analysis and safety-based design in safety-critical domains such as aviation [20], medical devices [5, 21], and automotive systems. Most previous studies used STPA only to derive the high-level safety constraints and identify the unsafe interactions that should be eliminated or controlled during the design process. However, here we further used the causal factors identified by the STPA analysis to identify the types and locations of faults to be injected into software to empirically assess the system's safety under realistic hazard scenarios.

## 7   Conclusions

This paper presents a framework for validating the robustness of safety mechanisms in robotic telesurgical systems. A systems-theoretic hazard analysis technique, STPA, was used to determine the safety hazard scenarios and their potential causes, in robotic surgical systems. A software-implemented fault injection framework was developed to simulate hazard scenarios by emulating the impact of intermittent and permanent faults in the robotic control software and hardware of the RAVEN II robot.

Software-implemented fault injection directed by the systems theoretic hazard analysis enables us to: (i) identify the safety hazard scenarios and determine their potential causes; (ii) trace propagation of faults in the system and discover the vulnerabilities in system safety mechanisms; (iii) determine strategic placement of new detectors that can mitigate the propagation of causal factors into safety hazards; (iv) provide useful feedback to the system developers on how to improve the safety mechanisms in the next-generation of devices. In particular, the identified hazard scenarios and the propagation paths from causal factors to safety hazards can be used for design of hazard prediction and mitigation mechanisms in the system. The proposed software fault-injection framework can be also used for simulating realistic safety-hazard scenarios experienced in the field during robotic surgical training.

**Acknowledgements.** A non-restricted grant from Infosys and a faculty award from IBM partially supported this work. Our special thanks to Blake Hannaford and researchers at the University of Washington Biorobotics Lab for access to a RAVEN II robot. We also thank Frances Baker and Carol Bosley for their editing of the paper.